\pdfoutput=1

\documentclass[journal]{IEEEtran}
\usepackage{cite}
\usepackage{amsmath,amssymb,amsfonts}
\usepackage{algorithmic}
\usepackage{graphicx}
\usepackage{textcomp}
\usepackage{graphics}
\usepackage{subfig}
\usepackage{cite}
\usepackage{booktabs}

\usepackage{multirow}

\newcommand{\tabincell}[2]{\begin{tabular}{@{}#1@{}}#2\end{tabular}}

\ifCLASSINFOpdf
\else
\fi
\begin{document}

\title{Partial Computing Offloading Assisted Cloud Point Registration in Multi-robot SLAM}

\author{Biwei Li,
        Zhenqiang Mi,~\IEEEmembership{Member,~IEEE,}
        Yu Guo,~\IEEEmembership{Student Member,~IEEE,}
        Yang Yang
        and
\\
       ~Mohammad S. Obaidat,~\IEEEmembership{Fellow,~IEEE}
\thanks{Manuscript received... This work has been supported by the National Natural Science Foundation of China under Grant No. 61772068, Grant 61472033 and Grant 61672178, Fundamental Research Funds for the Central Universities under Grant No. FRF-GF-17-B28, and Ministry of Education of the People's Republic of China Oversea Distinguished Professorship Grant given to Prof. Obaidat under Grant No. MS2017BJKJ003.  \emph{(Corresponding author: Zhenqiang Mi.)}}
\thanks{B. Li, Z. Mi, Y. Guo, and Y. Yang are with the School of Computer and Communication Engineering, University of Science and Technology
Beijing, Beijing 100083, China (e-mail: s20170639@xs.ustb.edu.cn; mizq@ustb.edu.cn; guoyu@xs.ustb.edu.cn; yyang@ustb.edu.cn;)}%
\thanks{M. S. Obaidat is with ECE Department, Nazarbayev University, Kazakhstan, King Abdullah II School of Information Technology, the University of Jordan, and PR China Oversees Distinguished Professor, University of Science and Technology Beijing, China. (e-mail: m.s.obaidat@ieee.org)}}


\maketitle

\begin{abstract}
Multi-robot visual simultaneous localization and mapping (SLAM) system is normally consisted of multiple mobile robots equipped with camera and/or other visual sensors. The networked robots work independently or cooperatively in an unknown scene in order to solve autonomous localization and mapping problem. One of the most critical issues in Multi-robot visual SLAM is the intensive computation that is normally required yet overwhelming for inexpensive mobile robots with limited on-board resources. To address this problem, a novel task offloading strategy and dense point cloud map construction method is proposed in this paper. First, we develop a novel strategy to remotely offload computation-intensive tasks to cloud center, so that the tasks that could not originally be achieved locally on the resource-limited robot systems become possible. Second, a modified iterative closest point algorithm (ICP), named fitness score hierarchical ICP algorithm (FS-HICP), is developed to accelerate point cloud registration. The correctness, efficiency, and scalability of the proposed strategy are evaluated with both theoretical analysis and experimental simulations. The results show that the proposed method can effectively reduce the energy consumption while increase the computation capability and speed of the multi-robot visual SLAM system, especially in indoor environment.
\end{abstract}


\IEEEpeerreviewmaketitle

\section{Introduction}

\IEEEPARstart{T}{he} aim of simultaneous localization and mapping (SLAM) technology is to provide good estimate of both the robot pose and the map in unknown environments, where an accurate map is unavailable\cite{Lee2015DV}. With the rapid development of onboard hardware, software and machine learning technology, increasing research interests have been witnessed on using SLAM technology to solve various complicated tasks in daily life, e.g., fire rescue, geological exploration, underwater search and archaeological excavation. In order to adapt to practical applications, one of the most notable research interests is three-dimensional SLAM. At present laser radar assisted camera is always used as signal inputs in 3D-SLAM solutions and commercially available products, such as drones. However, due to the high cost, fast power consumption and low precision of laser radar, the further application of the laser radar based 3D-SLAM technique is hindered to be widely used in the future.

3D \textbf{visual} SLAM, which uses depth cameras as the only sensor inputs, is considered to be beneficial in scenarios where the requirement of cost, energy, and weight of the system is strict. Due to various advanced cameras, 3D visual SLAM has attracted more and more attention and opened up a whole new range of possibilities in robot autonomous navigation field, e.g., virtual reality(VR), augmented reality(AR), and of cause, autonomous driving technology. However, 3D visual SLAM still faces the challenges of extensive amount of computation and communication tasks where powerful CPU and other on-board resources are required. It turns out that the cost of a single robot is too high to be applied to the actual scenario. Thus, a new performing model or strategy is of great necessity in the 3D visual SLAM process.

To overcome the aforementioned problem, more researches have been conducted to incorporate mobile multi-robot system into SLAM, so that the complexity of a task can be shared by a group of small, often less expensive mobile robots. Generally speaking, mobile robots are limited in size and power, which hinders them from carrying powerful computation and storage unit. Consequently, it might not be fair to ask them to perform extensive computation locally. In order to get rid of this bondage, the ¡°cloud robotics¡± proposed by Dr. Kuffner\cite{10031099795} in 2010 provides us with a new way to handle complex robot tasks. In cooperation with the cloud platform and data center, robotic network can improve its ability by a large margin. As for applying cloud robotics to solve multi-robot visual SLAM problems, there are still some challenges to be tackled.

We summarize the main problems faced by the multi-robot 3D visual SLAM into the following three aspects.

\par
\begin{enumerate}
\item The main task of 3D visual SLAM system is image processing, which requires intensive calculations. However, for most inexpensive robotic embedded devices (e.g., Raspberry Pi), it can hardly be handled in real time. Moreover, some high robustness visual SLAM systems, such as ORB-SLAM\cite{Mur2015ORB} and RTAP-MAP\cite{Labbe2014Online}, can only be calculated by powerful computer or even a server. Therefore, how to apply these visual SLAM systems to robotic network needs to be investigated.
\par
\item In order to solve the problem of excessive calculations, some researchers have proposed using cloud robotics to offload tasks to cloud. However, images are the only data for visual SLAM system. The memory space occupied by image is very large, which may cause certain problems while transmitting, such as communication loss and unexpected latency. For example, Wi-Fi connections are not invariably usable and reliable\cite{Ning2017Social}. Moreover, security issues such as hacker interception tend to occur when using images as the only input. Hence, a safe and efficient offloading strategy is required.
\par
\item The state of art SLAM systems which can be used in robot embedded devices always have latency and cannot make sure the robustness of the system. After a long running, robots always lost themselves and cannot be restarted in a short time. In the condition of absence of any priori artificial calibration, how to achieve real-time calculation as well as ensure system robustness also need to be studied.
\end{enumerate}
\par
According to the above analyses, it is found that the main problems need to be solved are the insufficient on-board computing power and limited network transmission bandwidth, which cause multi-robot visual SLAM system very hard to achieve high robustness and real-time calculation. Therefore, we will tackle the aforementioned challenges with a two-step strategy. In the first step, we elaborate on the partial computing offloading method. Second, we proposed a novel iterative closest point algorithm to reduce the computational complexity of multi-robot visual SLAM system.

The main contributions of our work can be summarized as follows.

\par
\begin{enumerate}
\item \textit{A novel partial offloading strategy:} It is built to balance the calculation of local robots and cloud. To take full advantage of local computing and cloud computing, a strategy to offload most of the computation into the powerful cloud while reducing the amount of data which needs to be transmitted is designed. An efficient algorithm is proposed to find the best offloading point of the system. By this way, incomplete information is offloaded to the cloud in which case even if a hacker intercepts the network, it is hard for them to steal real and complete image information. In addition, this method greatly reduces the energy consumption as well as decreases the processing time.
\par
\item \textit{Improvement of point cloud map registration algorithm:} One of the two main tasks of SLAM is map building, which is especially important for multi-robot SLAM system. The robots work independently in the environment and the maps built by each of them need to be matched. In order to reduce the complexity of the whole SLAM system, we improved classic registration algorithm, iterative closest point (ICP) algorithm, by proposing a novel fitness score hierarchical iterative closest point (FS-HICP) algorithm. The improved algorithm reduces the time costs and energy consuming.
\item \textit{Construction of dense and semi-dense point cloud map:} We build global dense point cloud map because it is more universal and observable. Additionally, due to the fact that multi-robot may be used for obstacle avoidance, path planning and other applications. In order to describe whether the scene is accessible, a semi-dense octree map is also constructed.
\par
\end{enumerate}
\par
We also conduct extensive simulation and experimental evaluations to demonstrate the efficiency of the methods. Our partial offloading strategy and the FS-HICP algorithm can greatly reduce the whole system running time and save the energy consuming.

The rest of this paper is organized as follows. Section II presents a brief survey of the related works. Section III presents the system model, and then introduces the formulation of the partial offloading problem and multi-robot registration model. Details of our method are shown in Section IV. Section V gives the evaluation results. Finally, Section VI concludes the paper and presents our proposed future work.

\section{RELATED WORK}
Visual SLAM, as an effective and economical way to help robot localization and mapping in an unknown environment, has reached substantial robustness and accuracy in the centimeter range for single robot applications\cite{Mur2016ORB, 8606275}. ORB-SLAM2\cite{Mur2016ORB} system proposed by Raul Mur-Artal et al., which uses three threads to work simultaneously, greatly improves the efficiency of the algorithm. As multi-robot systems have been becoming popular in numerous scenarios, ranging from search and rescue applications to digitization of archeological sites. Some researchers\cite{Koch2015Multi, Shamsudin2018Consistent, Sasaoka2016Multi, Indelman2014Multi} aim to improve the multi-robot SLAM network, for example \cite{Sasaoka2016Multi} uses an information fusion technology, in which the information processed by the single robot is collected and optimized by the base station. Another example \cite{Indelman2014Multi} is based on the multi-robot altitude map positioning of the unknown initial relative pose. Other researchers focus on improving the speed and efficiency of the whole system, for instance, MTM (map transformation matrix) is a multi-robot fast SLAM proposed by HeonCheol Lee\cite{HeonCheol2012Probabilistic}, which only calibrates the particle of the nearest point. Of course, there are a lot of successful multi-robot robot models\cite{Lee2017Iterative, Fabresse2018An, Isobe2018Occlusion, Nowicki2017An}, such as five-legged robot\cite{Nowicki2017An}, using SLAM technology to achieve different kinds of tasks in unknown environment.

Because the applications environment of multi-robot SLAM are relatively complex, in order to solve the calculation problem, some successful framework based on cloud robot architecture have been proposed, such as DaVinci\cite{Arumugam2010DAvinCi}, which is a software framework that provides the scalability and parallelism advantages of cloud computing implemented in the FastSLAM algorithm. Rapyuta\cite{Hunziker2013Rapyuta}, an open source Platform-as-a-Service (PaaS) framework designed specifically for robotics applications such as SLAM, can help robots to offload heavy computation by providing secured customizable computing environments in the cloud. All these frameworks have verified the feasibility and effectiveness of using ``cloud + robot" synergy in solving SLAM problems.

The performance of robots depends not only on their motion mechanism, but also on their familiarity with the environment. At present, the two-dimensional map creation technology has been well developed, and the three-dimensional map creation technology still has great potential in the enhancement of efficiency and reduction of cost. A tailoring and scaling iterative closest point (TsICP) algorithm was proposed by Liang Ma\cite{Ma2016Merging}, which considers the problem of merging grid maps with different resolutions. Authors in \cite{Yang2016Go} introduce a global optimal solution to Euclidean registration in 3D. The work in \cite{Marani2016A} modifies the well-known iterative closest point (ICP) algorithm by introducing the concept of deletion mask. Except using traditional method, Rebro University's Magnusson team proposed a 3D NDT point cloud registration method\cite{Magnusson2007Scan}, which has little correlation with the initial value. Since then, many algorithms have been improved based on NDT algorithm. In terms of application, Kinect Fusion\cite{Newcombe2012KinectFusion} proposed by Imperial College and Microsoft Research Institute in 2011, using the point-plane method to realize point cloud map registration, which opened a prelude of real-time 3D reconstruction with RGB-D camera.

To date, most successful local SLAM systems have the problem of too much calculation. Thus, the ``cloud robotic" technology is proposed to solve this problem. However, they mainly focus on the multi-robot network structure, network transmission path, data compression and so on, without considering combining cloud computing and local computing, such as \cite{Torkashvan2012CSLAM, Benavidez2015Cloud, 8456343, 8620865}. Different from most works mentioned above, in this paper, a novel strategy is used to choose a offloading point, reducing energy consuming by offloading part of the calculation to cloud. In addition, this paper proposes an improved ICP algorithm realizing multi point cloud map registration, with the objective to reduce the times of iterations as well as improve convergence accuracy.

\section{SYSTEM MODEL AND PROBLEM FORMULATION}
In this section, firstly the total system framework is presented. Then the model of multi-robot partial computing offloading is proposed. Finally, the multi-robot dense point cloud map model and its calculation method is introduced.

\subsection{System Framework}
In this paper, it is assumed that multi-robot system works independently to execute tasks of SLAM in an unknown indoor scenario. In the SLAM system, the two main tasks are localization and mapping. Localization which includes tracking thread, localmapping thread and loopclosing thread is mainly meant to calculate the trajectory and keyframe of the robot in an unknown environment. Mapping is mainly used to construct a global map of the scene based on local maps got by each robot. In the system model, the keyframes and trajectory got by localization of each robots are used by the mapping process.

During the localization process, robots may lost themselves unexpectedly, especially when running for an extended period of time because the indoor environment is usually complicated. If the complexity of the algorithm for better robustness and accuracy is increased, it will lead to too much computation. On this condition if all the data (captured images) is offloaded to cloud to compute, the transmission will be too slow due to large amount of volumes. Meanwhile, during the mapping process, the partial 3-D dense map built by each robot takes up too much memory. In order to unify them into the global coordinate, very large amount of calculation is needed, which will cost a lot of time and energy. Therefore, in this paper the partial computing offloading model and improved registration of point cloud map model are used to solve the problems mentioned above. The most suitable point is researched to choose to offload parts of the tasks, which require extensive calculation, to cloud in order to reduce the total energy consumption of the localization process. Furthermore, a novel point cloud registration algorithm is utilized to reduce the time of dense map building in the mapping process.

\begin{figure}[tbp]
  \centering
  \includegraphics[width=3.5in]{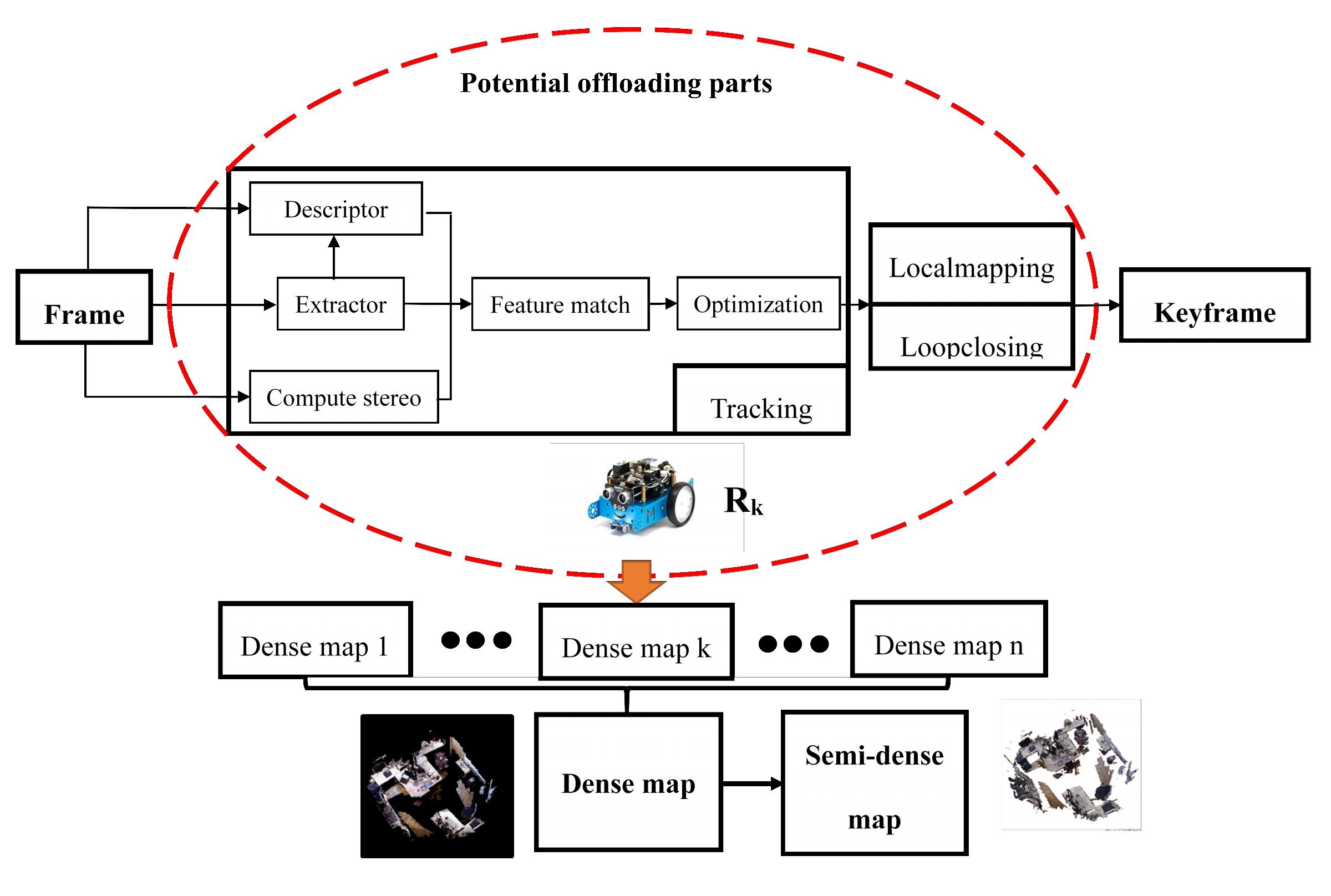}
  \caption{Overview of the system}
  \label{fig31}
\end{figure}

It is assumed that there are several heterogeneous robots $R=\{R_{1},R_{2},...R_{n}\}$ which build local maps from different starting points. Each robot gets images\textit{(Frames)} independently. During the localization process, the most suitable offloading point is chosen to offload some tasks to cloud for further calculating. The keyframe and trajectory obtained by each robot will be used to build partial dense point cloud map, which can be expressed as $Dense map=\{Dense map_{1},Dense map_{2},...Dense map_{n}\}$. Finally, during the mapping process, the partial maps match one by one to get a global coordinate with our novel point cloud registration algorithm. The overall framework is shown in Fig. \ref{fig31}.

The whole system energy consumption of our framework is noted as $E_{system}$, the energy consumption of localization process as $E_{localization}$ while $E_{mapping}$ represents the energy consumption of mapping process. Hence, the $E_{system}$ can be expressed in (\ref{eq50}).
\begin{equation} \label{eq50}
E_{system}=E_{localization}+E_{mapping}
\end{equation}
\par

The basic viewpoint in our problem is that the smaller the value of $E_{system}$ which can be seen in (\ref{eq51}), the better the system works. So the problem can be divided into two seperate parts: to get the smaller localization part and mapping part as far as possible.
\begin{equation} \label{eq51}
\begin{split}
min(E_{system})=min(E_{localization}+E_{mapping})\\
=min(E_{localization})+min(E_{mapping})
\end{split}
\end{equation}
\par

The next two parts will introduce the models to get the minimum value of $E_{localization}$ and $E_{mapping}$, respectively.

\subsection{Partial Computing Offloading Model}
\begin{figure}[htbp]
  \centering
  \includegraphics[width=3.5in]{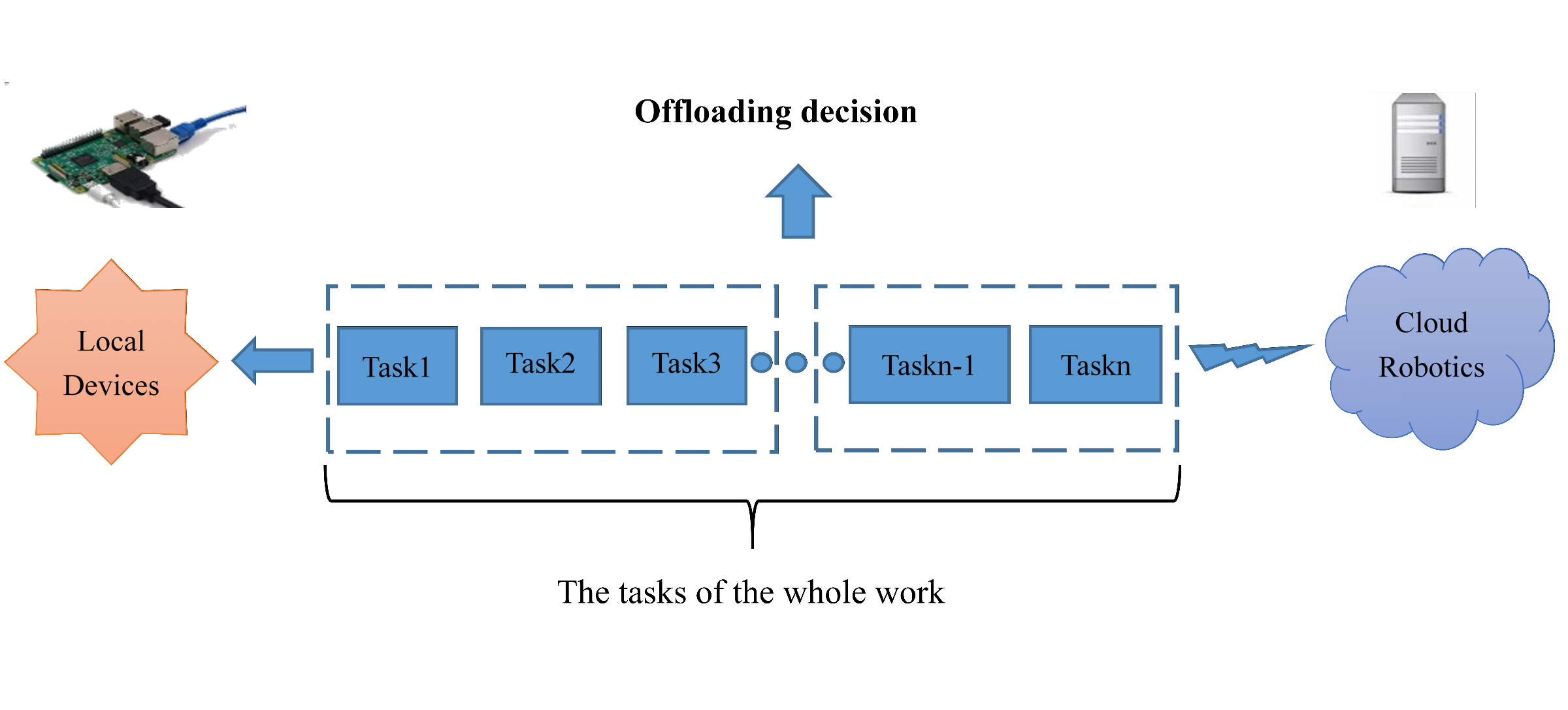}
  \caption{Partial offloading model}
  \label{fig33}
\end{figure}

In a nutshell, partial computing offloading can be summarized as encapsulating partial high-density computing tasks that would otherwise be performed by local robot nodes into a central processor or cloud to execute, by which way the energy consumption of robot can be reduced and the network efficiency will be improved. It is shown in Fig. \ref{fig33}.

In the multi-robot system, suppose the energy consumption of robot $R_{k}$ is $E_{lk}$, $E_{localization}$ can be summarized as (\ref{eq52}).
\begin{equation} \label{eq52}
E_{localization}=\sum_{k=0}^{n}E_{lk}
\end{equation}
\par
For each robot in the system has the same status level; we can calculate the minimum value of each $E_{lk}$ in order to get the minimum value of the $E_{localization}$,. In the partial offloading model proposed, some parts of the tasks are calculated by local devices while the others are calculated by cloud. It is supposed that $E_{local}$ represents the energy consumed by one of the local embedded device while $E_{cloud}$ represents the energy consumed by cloud. Transmission consuming is expressed by $E_{tr}$. The calculation of $E_{lk}$ is given in (\ref{eq53}).

\begin{equation} \label{eq53}
E_{lk}=E_{local}+E_{cloud}+E_{tr}
\end{equation}
\par

\textit{Problem description:} With $E_{lk}$ as the energy consumption of each robots, our aim is to choose a suitable offloading point, which can make each $E_{lk}$ reach its minimum value, at which time the $E_{localization}$ can reach its minimum value.

\subsection{Registration Of Point Cloud Map Model}
If every frame of the point cloud is merged into the map, the capacity of the local map will be too large, which will reduce the real-time performance of the system. As the pose of adjacent frames change very little, it is proposed to use key frames to build the point could map.

\begin{figure}[htbp]
  \centering
  \includegraphics[width=3.5in]{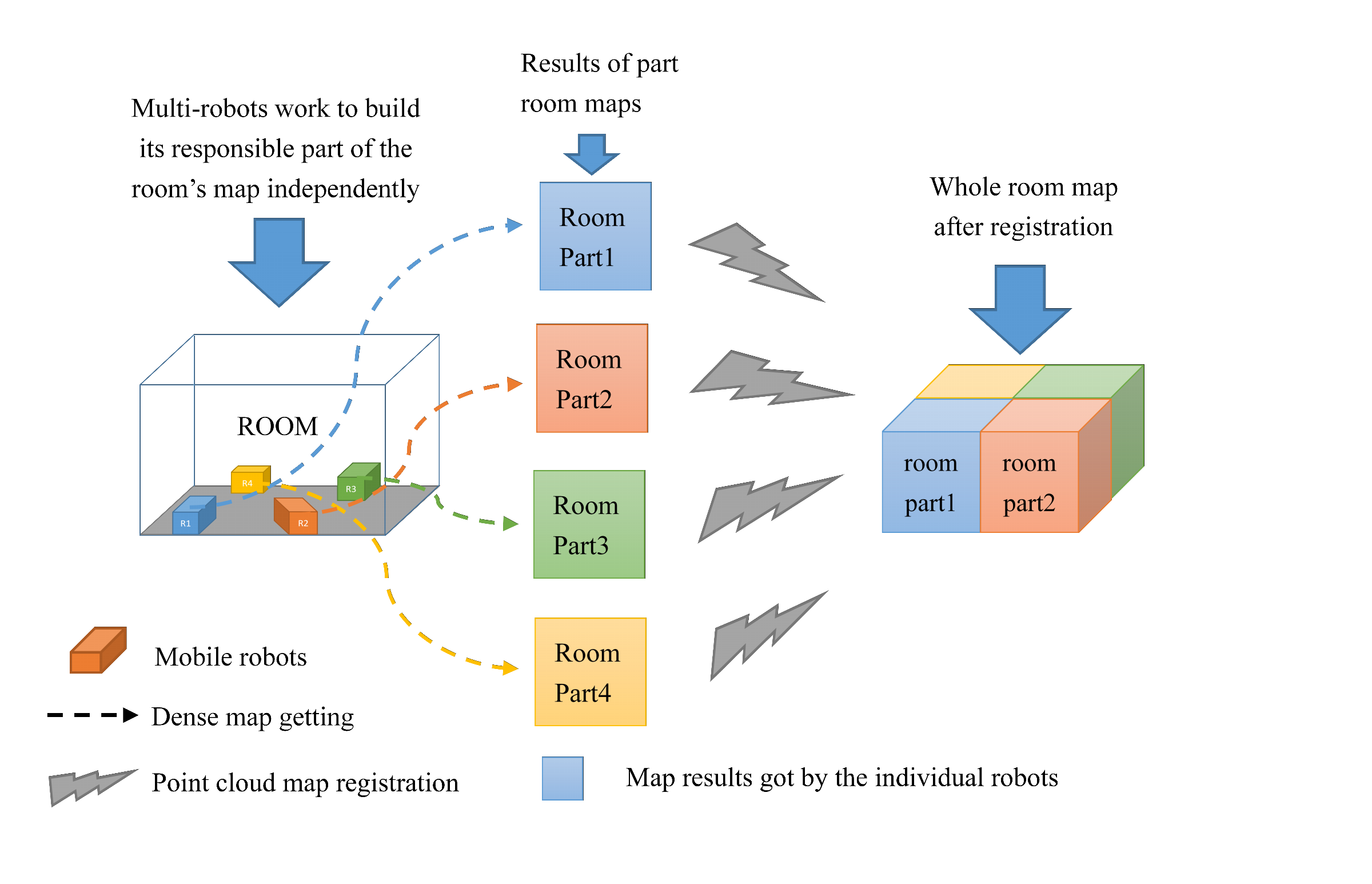}
  \caption{Multi-robot dense map registration model}
  \label{fig32}
\end{figure}

It is supposed that we have a multi-robot $R=\{R_{1},R_{2},...R_{n}\}$ moves in a room. The partial dense point cloud maps of the room built by these robots can be expressed as $Room=\{Room_{1},Room_{2},...Room_{n}\}$. These maps need fusing one by one and unifying to a global coordinate and finally a complete global point cloud map of the whole room is built as shown in Fig. \ref{fig32}.

The energy consumption of mapping process is expressed by $E_{mapping}$. In this paper, three-dimensional global dense map is built in cloud. If the power of cloud is $P_{cloud}$ and the time used by mapping processing is $T_{mapping}$, the $E_{mapping}$ can be expressed as (\ref{eq54}).
\begin{equation} \label{eq54}
E_{mapping}=P_{cloud}\cdot T_{mapping}
\end{equation}

\textit{Problem description:} Our aim is to propose a novel point cloud registration method in order to make the value of $T_{mapping}$ and $E_{mapping}$ as small as possible.

\section{Method Description}
The objective of this section is to introduce our strategy to find the minimum value of $E_{lk}$ stated in section III, and then introduce an improved point cloud map registration algorithm, Fitness Score Hierarchical Iteration Closest Point(FS-HICP) algorithm, in order to reduce the value of $E_{mapping}$.
\subsection{Offloading Point Selection}
Generally speaking, mobile robots use embedded devices as processing unit. It is supposed that the power of the local embedded devices, in watts, is $P_{local}$, while $P_{cloud}$  represents the power of cloud server. $C_{local}$ and $C_{cloud}$ are used to denote the data quantity to be calculated by local devices and cloud. $U_{local}$ and $U_{cloud}$, in GB per second, represent the data processing speed of the local devices and the cloud, respectively. Calculations of $E_{local}$ and $E_{cloud}$ mentioned in section III are shown in (\ref{eq10}) and (\ref{eq11}).

\begin{equation} \label{eq10}
E_{local}=P_{local}\cdot \frac{C_{local}}{U_{local}}
\end{equation}
\par

\begin{equation} \label{eq11}
E_{cloud}=P_{cloud}\cdot \frac{C_{cloud}}{U_{cloud}}
\end{equation}
\par

It is assumed that $P_{tr}$ is the power for sending and receiving data. $T_{tr}$ represents the time consuming of the transmission processing. The calculation of $E_{tr}$ is shown in (\ref{eq12}).
\begin{equation} \label{eq12}
E_{tr}=P_{tr}\cdot T_{tr}
\end{equation}
\par

So the $E_{lk}$ can be summarized as in (\ref{eq13}).
\begin{equation} \label{eq13}
\begin{split}
E_{lk}=E_{local}+E_{cloud}+E_{tr}=\\
P_{local}\cdot \frac{C_{local}}{U_{local}}+P_{cloud}\cdot \frac{C_{cloud}}{U_{cloud}}+P_{tr}\cdot T_{tr}
\end{split}
\end{equation}
\par

Suppose $\eta $ is the offloading point. The time consumed by tasks which have been executed divided by the time consumed by total tasks as the value of $\eta$, which can be seen in (\ref{eq31}). The value of $\eta$ is between 0 and 1. When $\eta=0$, all the tasks are computed by cloud and when $\eta=1$, all the tasks are computed by local embedded devices. So, $C_{local} $ and $C_{cloud} $ can be rewritten as is shown in equation (\ref{eq14}).
\begin{equation} \label{eq31}
\eta=\frac{\sum_{i=0}^{k}T_{taski}}{\sum_{j=0}^{n}T_{taskj}}\; (0\leq k\leq n)
\end{equation}
\par

\begin{equation} \label{eq14}
\left\{\begin{matrix}
C_{local}=C_{total}\cdot \eta \\
C_{cloud}=C_{total}\cdot (1-\eta )
\end{matrix}\right.
\end{equation}
\par

Because the transmission time is related to the amount of data that needs to be transferred at the offloading point, $T_{tr}$ is a function of $\eta $ as shown in (\ref{eq15}).

\begin{equation} \label{eq15}
T_{tr}=f(\eta )
\end{equation}
\par

Thus, we substitute (\ref{eq14}) and (\ref{eq15}) into (\ref{eq13}) to get (\ref{eq16}).
\begin{equation} \label{eq16}
\begin{split}
E_{lk}=P_{local}\cdot \frac{C_{total}\cdot \eta }{U_{local}}+\\
P_{cloud}\cdot \frac{C_{total}\cdot (1-\eta )}{U_{cloud}}+P_{tr}\cdot T_{tr}(\eta )
\end{split}
\end{equation}
\par

From (\ref{eq16}) we known, the value of $\eta$ determines the value of $E_{lk}$. In our experiment parts, we will choose a suitable offloading point $\eta$ in order to make $E_{lk}$ get its minimum value. When we know the best offloading point, the steps of partial computing offloading model can be seen as follow:
\begin{enumerate}
\item Get frames from Kinect camera.
\par
\item Based on the value of $\eta$, some tasks are calculated on local embedded devices.
\par
\item Store the results calculated by local embedded devices in a YAML file.
\par
\item Send the message(YAML file) to cloud by ROS system.
\par
\item Calculate the other part of the SLAM on cloud using the data uploaded by local devices.
\end{enumerate}

\begin{figure}[hbtp]
  \centering
  \includegraphics[width=9cm]{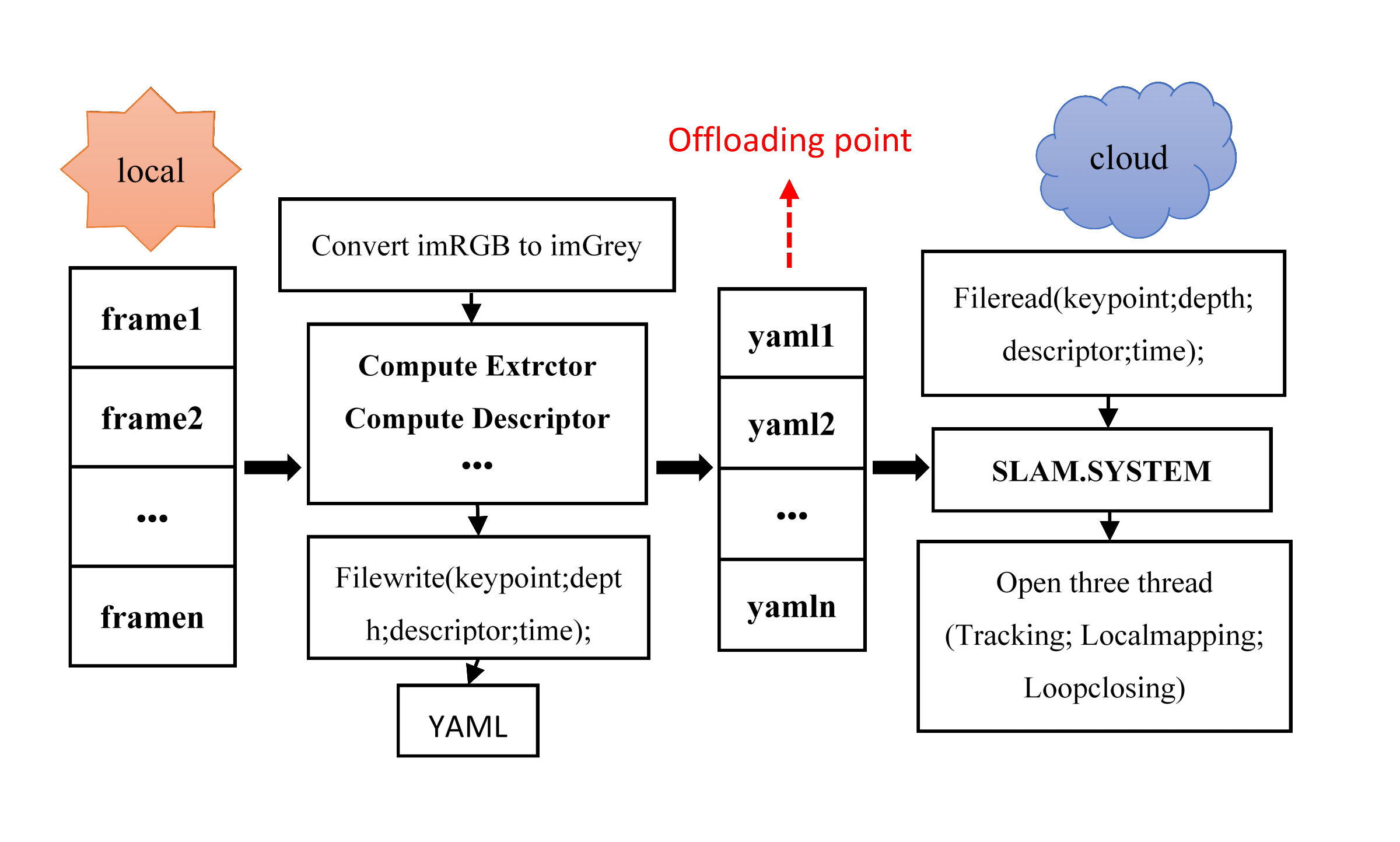}
  \caption{Partial offloading process}
  \label{fig3}
\end{figure}

The specific implementation process is shown in Fig. \ref{fig3}.

The main notations in the offloading model are listed in Table \ref{tab11}.
\begin{table}[hbtp]
\caption{Important notations}
\centering
\label{table}

\begin{tabular}{|c|c|}
\hline
Notation&
Description\\
\hline
$P_{local}$&
Power of local devices\\
\hline
$P_{cloud}$&
Power of cloud\\
\hline
$P_{tr}$&
Power of network transmission\\
\hline
$U_{local}$&
Data processing speed of local devices \\
\hline
$U_{cloud}$&
Data processing speed of cloud robotics\\
\hline
$C_{local}$&
Calculation of local devices\\
\hline
$C_{cloud}$&
Calculation of cloud\\
\hline
$C_{total}$&
Total calculation amount\\
\hline
$T_{tr}$&
Transmission time\\
\hline
$\eta $&
Offloading point(Independent variable)\\
\hline
\end{tabular}
\label{tab11}
\end{table}

\subsection{FS-HICP Algorithm}
Suppose $Room_{p}$ and $Room_{q}$ are the two point cloud data sets needed to be matched and we simply mark them as $P$ and $Q$. First, we search each point $p_{i}$ in the point set $P$ for its nearest point $q_{i}$ in the point set $Q$ as the corresponding point. Then the corresponding set of $P=\{p_{1},p_{2},...p_{n}\}$ is set to $Q=\{q_{1},q_{2},...q_{n}\}$ to solve a European transformation $R$, $t$ in order to satisfy (\ref{eq0}).
\begin{equation} \label{eq0}
\forall i, p_{i}=Rq_{i}+t
\end{equation}

The centroid position $p$ and $q$ of the two sets of points is calculated and so are the coordinates of each point after removing the centroid, shown in (\ref{eq1}).
\begin{equation} \label{eq1}
{p_{i}}'=p_{i}-p, {q_{i}}'=q_{i}-q
\end{equation}

Then the rotation matrix based on the following optimization formula is calculated (\ref{eq2}).
\begin{equation} \label{eq2}
R^{*}=arg\min \frac{1}{2}\sum_{i=1}^{n}||{p_{i}}'-R{q_{i}}'||^{2}
\end{equation}

Finally, the $t$ is derived according to $R$, as shown in (\ref{eq3}).
\begin{equation} \label{eq3}
t^{*}=p-Rq
\end{equation}

The transformation matrix is a fourth-order matrix contains translation($t$) and rotation($R$), which can be expressed as (\ref{eq30}).
\begin{equation} \label{eq30}
T=\begin{bmatrix}
R &t \\
0& 1
\end{bmatrix}
\end{equation}

The classical ICP algorithm(\textit{point-point} algorithm) proposed by Besl and McKay\cite{Besl1992Method} can be described as $P$ is unchangeable while $Q$ rotates and translates iteratively to get as close as possible to the $P$. In the improvements of the classic algorithm, the most famous one is the \textit{point-plane} method used by Kinect Fusion\cite{Newcombe2012KinectFusion}. However, both of them cannot achieve registration, quickly and the iteration times and the overall time costs are too much.

In this paper, the hierarchical iterative closest point algorithm based on fitness score (FS-HICP algorithm) is proposed. There are three main improvement in this novel algorithm.

\begin{enumerate}
\item The stop condition according to the change of fitness score is carefully revised.

\par
\item The idea of layered filtering is used to accelerate iteration.
\par
\item In the iterative process, the maximum response distance will be adjusted according to the matching quality, so that the algorithm can converge in a relatively short time.

\end{enumerate}

\begin{figure}[htbp]
  \centering
  \includegraphics[width=3.3in]{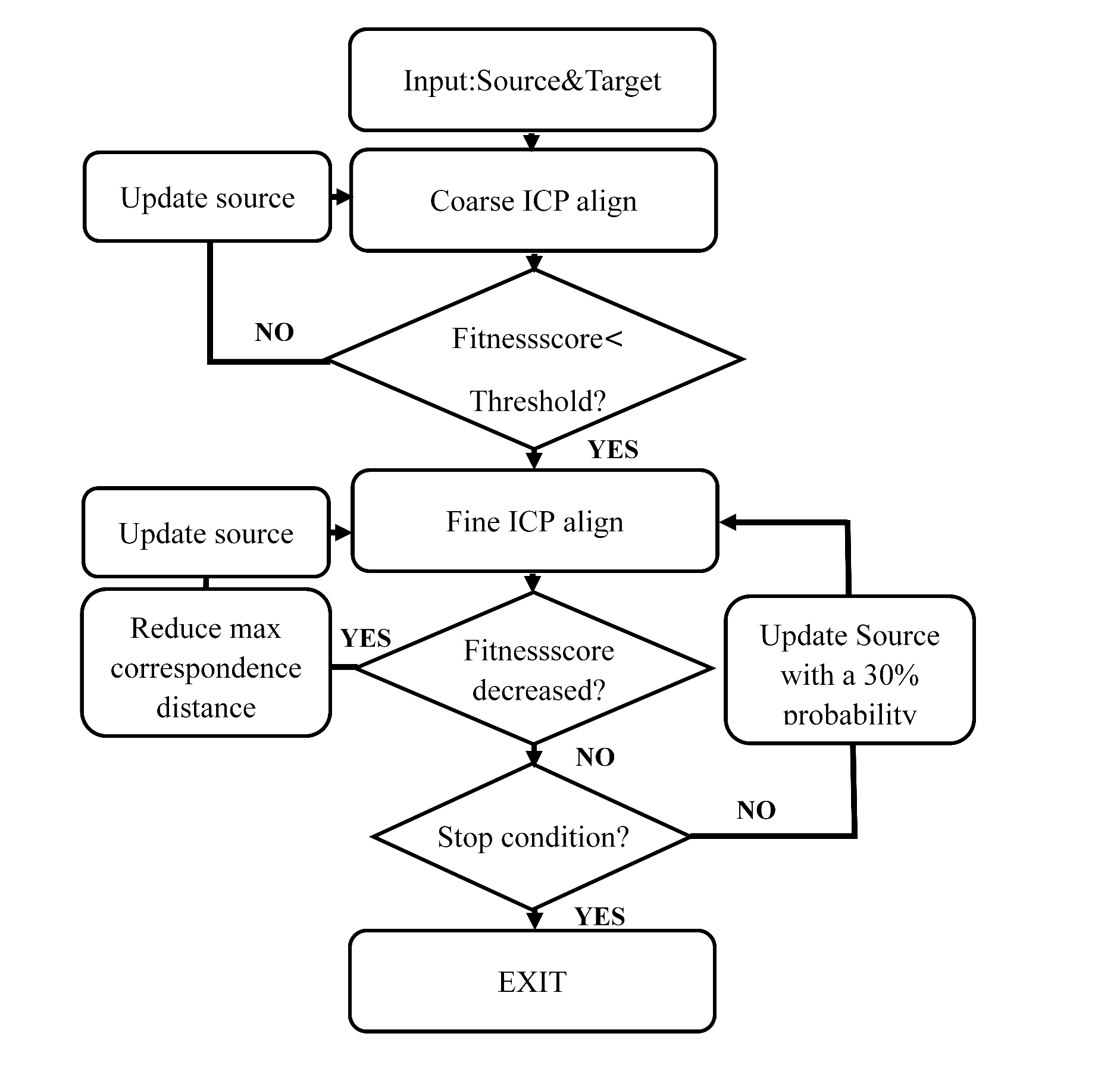}
  \caption{The flow of the FS-HICP algorithm}
  \label{fig5}
\end{figure}

The flow of the FS-HICP algorithm is shown in Fig.\ref{fig5}.

The specific execution steps of the algorithm are as follows:

\begin{enumerate}
\item Delete the error point and useless point. In this paper, a point with a depth value of 0 or a depth value greater than 7000 is considered to be an invalid point and it will be rejected. Then the points are filtered and the voxel filter size is 0.1.
\par
\item Match two point cloud map. In the process the algorithm is iterated twice internally.
\par
\item Determine whether the fitness score is less than a given threshold (50\%). If not, the source is updated and return to the second step and continue iteration.
\par
\item Filter with a voxel filter size of 0.05.
\par
\item Match two point cloud map. In the process the algorithm is iterated twice internally.
\par
\item Determine whether the fitness score reduces. If not, the source is updated with a certain probability (30\%) and return to the fifth step to continue searching the iteration direction. If the fitness score doesn't reduce more than three times or the algorithm reaches the convergence precision, the iteration is stopped. If the fitness score decreases, reduce the minimum iteration distance between two points, update the source and return the fifth step to continue the iteration.
\end{enumerate}

The pseudo code of FS-HICP algorithm is shown in Table \ref{tab2}.

\begin{table}[htbp]
\caption{Pseudo code of the FS-HICP algorithm}
\label{table}

\begin{tabular}{p{230pt}}
\toprule[1.5pt]
\textbf{Input:}  Multiple point cloud maps:Data \par
\textbf{Output:} Transformation matrix:FinalTrans\\
\qquad \quad \ Global point cloud map\\
\hline
1.  \textbf{if} \ $(data==\o )$ \par
2.  \qquad \textbf{return} $ERROR$\par
3.  \textbf{end if}\par
4.  \textbf{for}$(i=1;i \le data.size;++i)$\par
5. \qquad $Source=data[i-1]$\par
6.  \qquad $Target=data[i]$\par
7. \qquad $Setvoxelgrid=0.1$\par
8. \qquad  $Setmaxcorrespondancedistance=0.2$\par
9. \qquad $Score=fitnessscore$\par
10. \qquad \textbf{while} $fitnessscore \ge 0.5*score$\par
11. \qquad \qquad  $Align(source)$\par
12. \qquad \qquad  $Source=source*trans$\par
13. \qquad \qquad  $FinalTrans= FinalTrans *trans$\par
14.  \qquad      $Setvoxelgrid=0.05$\par
15.  \qquad      $Setmaxcorrespondancedistance=0.1$\par
16.  \qquad     \textbf{do}\par
17. \qquad \qquad  $Align(source)$\par
18. \qquad \qquad  \textbf{if} fitnessscore decrease\par
19. \qquad \qquad \qquad $Setmaxcorrespondancedistance=$ \par
      \qquad \qquad \qquad \qquad  $maxcorrespondancedistance-0.001$\par
20. \qquad \qquad \qquad $Source=source*trans$\par
21. \qquad \qquad \qquad $FinalTrans= FinalTrans *trans$\par
22. \qquad \qquad \textbf{else}\par
23.  \qquad \qquad \qquad Update source with a 30\% probability\par
24.  \qquad \qquad  \textbf{end if}\par
25.   \qquad  \textbf{until} fitnessscore has not been reduced for 3 consecutive times.\par
26.  \textbf{return} $FinalTrans$\\

\bottomrule[1.5pt]
\end{tabular}
\label{tab2}
\end{table}

\section{EXPERIMENTAL RESULTS}
This section presents the evaluation results of the improved methods through extensive simulation and experimental studies. The results are two-fold: computing offloading and point cloud map fusion.

The embedded device used on the multi-robot system in this paper is Raspberry Pi and data is transferred by Wi-Fi network. The RGB-D image acquisition device is Kinect and the master node used in the simulation experiment is a desktop computer with 16G memory and 4-core 3.3 GHz CPU. The ROS system is used for network transmission. In the simulation experiments, the data sets used are RGB-D data sets from TUM (https://vision.in.tum.de/data/datasets/rgbd-dataset/download).

\subsection{Computing Offloading}
In this section, it is assumed that the total calculation is 1G. The values of the parameters in the partial computing offloading model are shown in Table \ref{tab10}\cite{5445167}. So the formula (\ref{eq16}) can be written as in (\ref{eq21}).

\begin{table}
\caption{Parameter values}
\centering
\label{table}

\begin{tabular}{|c|c|c|}
\hline
Parameters& Description&
Value\\
\hline
$P_{local}$& Power of local devices&
$0.9W$\\
\hline
$P_{cloud}$& Power of cloud&
$0.3W$\\
\hline
$P_{tr}$&  Power of network transmission&
$1.3W$\\
\hline
$U_{local}$& Data processing speed of local devices&
$1.2G/s$\\
\hline
$U_{cloud}$& Data processing speed of cloud&
$3.3G/s$\\
\hline
$C_{total}$&  Total calculation amount&
$1G$\\
\hline
\end{tabular}
\label{tab10}
\end{table}

\begin{equation} \label{eq21}
E_{lk}=\frac{0.9W}{1.2G/s}\cdot1G\cdot  \eta +\frac{0.3W}{3.3G/s}\cdot1G\cdot  (1-\eta )+1.3W\cdot T_{tr}(\eta )
\end{equation}

(\ref{eq21}) can be simplified to (\ref{eq22}).
\begin{equation} \label{eq22}
E_{lk}=0.659\cdot \eta +1.3T_{tr}(\eta )+0.091
\end{equation}

\begin{figure}[htbp]
  \centering
  \includegraphics[width=3in]{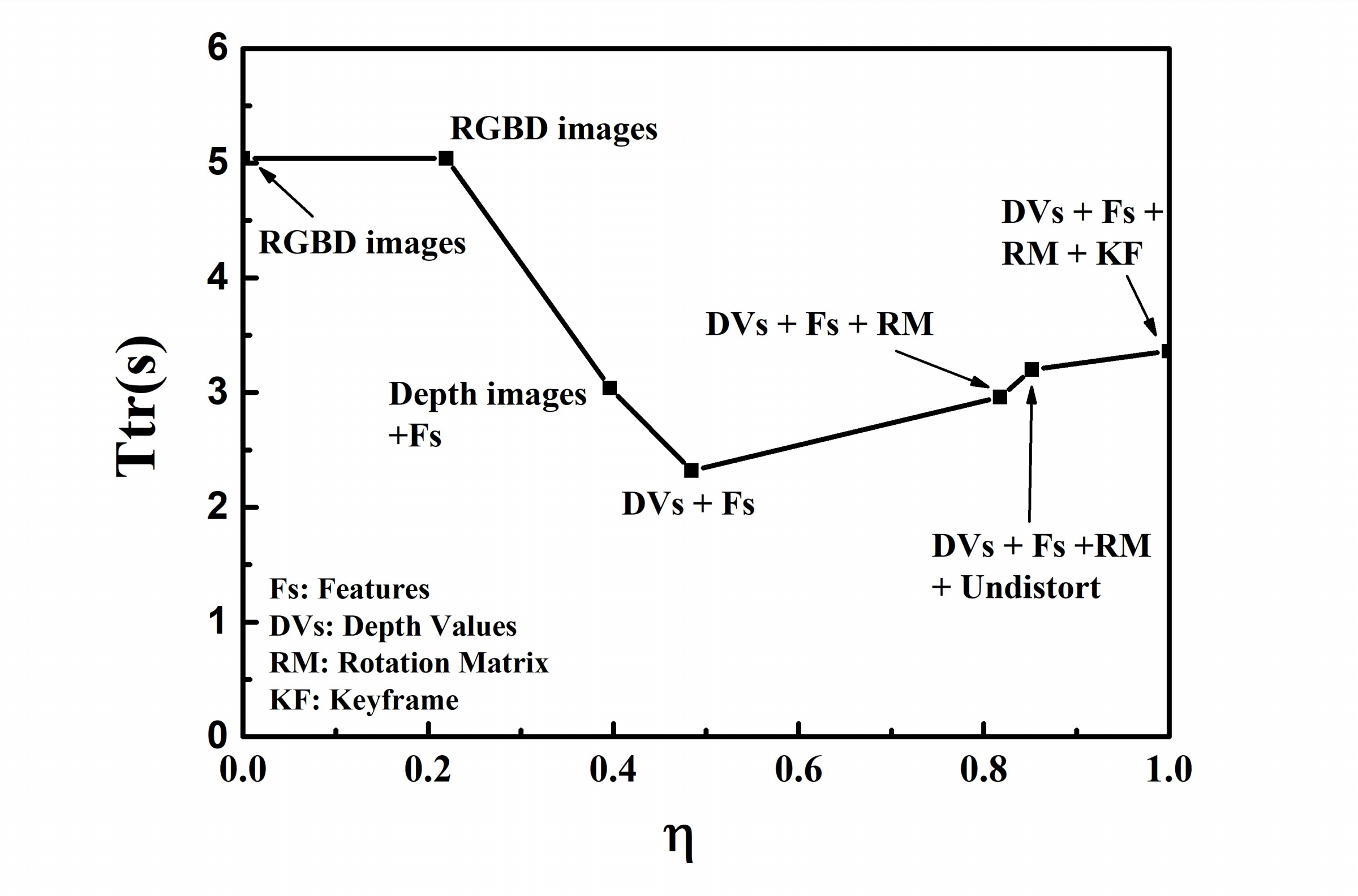}
  \caption{Transmission time}
  \label{fig34}
\end{figure}

During the localization process, there are several potential offloading points. As for processing 10 images, of which the pixel is $640\times 480$, the average time consumed by each part can be seen in Table \ref{tab50}. Based on (\ref{eq31}), we can calculate the value of $\eta$. Then according to our experiment, the minimum data needed to be offloaded at each possible offloading point and the memory occupied by these data can be described in Table \ref{tab15}. Suppose the network transmission rate is 10Mbps, the value of $T_{tr}$ is shown in Fig. \ref{fig34}.

\begin{table*}[tbp]

\caption{Time using of each part in the localization processing}
\label{table}
\centering
\begin{tabular}{|c|c|c|c|c|c|c|c|}
\hline
Items&
Extractor&
Descriptor&
Compute
stereo&
Feature match&
PnP&
Others&
Total\\
\hline
time(s)&
0.0154&
0.0124&
0.0062&
0.0236&
0.0022&
0.011&
0.0702\\
\hline
\end{tabular}
\label{tab50}
\end{table*}

\begin{table}
\caption{Data to be offloaded at each possible offloading point }
\centering
\label{table}

\begin{tabular}{|c|c|c|}
\hline
$\eta $& Data&
Memory footprint(M)\\
\hline
0& RGBimage+Depth image & 6.3\\
\hline
0.22& RGBimage+Depth image & 6.3\\
\hline
0.396& features +depth image & 3.8\\
\hline
0.484& features+depth value & 2.9\\
\hline
0.818& features+depth value+Rotation matrix & 3.7\\
\hline
0.852& \tabincell{c}{features+depth value+Rotation matrix \\+undistort} & 4.0\\
\hline
$\eta\rightarrow 1$ & \tabincell{c}{features+depth value+Rotation matrix \\ +undistort+keyframe} & 4.2\\
\hline
\end{tabular}
\label{tab15}
\end{table}

\begin{figure}[htbp]
  \centering
  \includegraphics[width=3in]{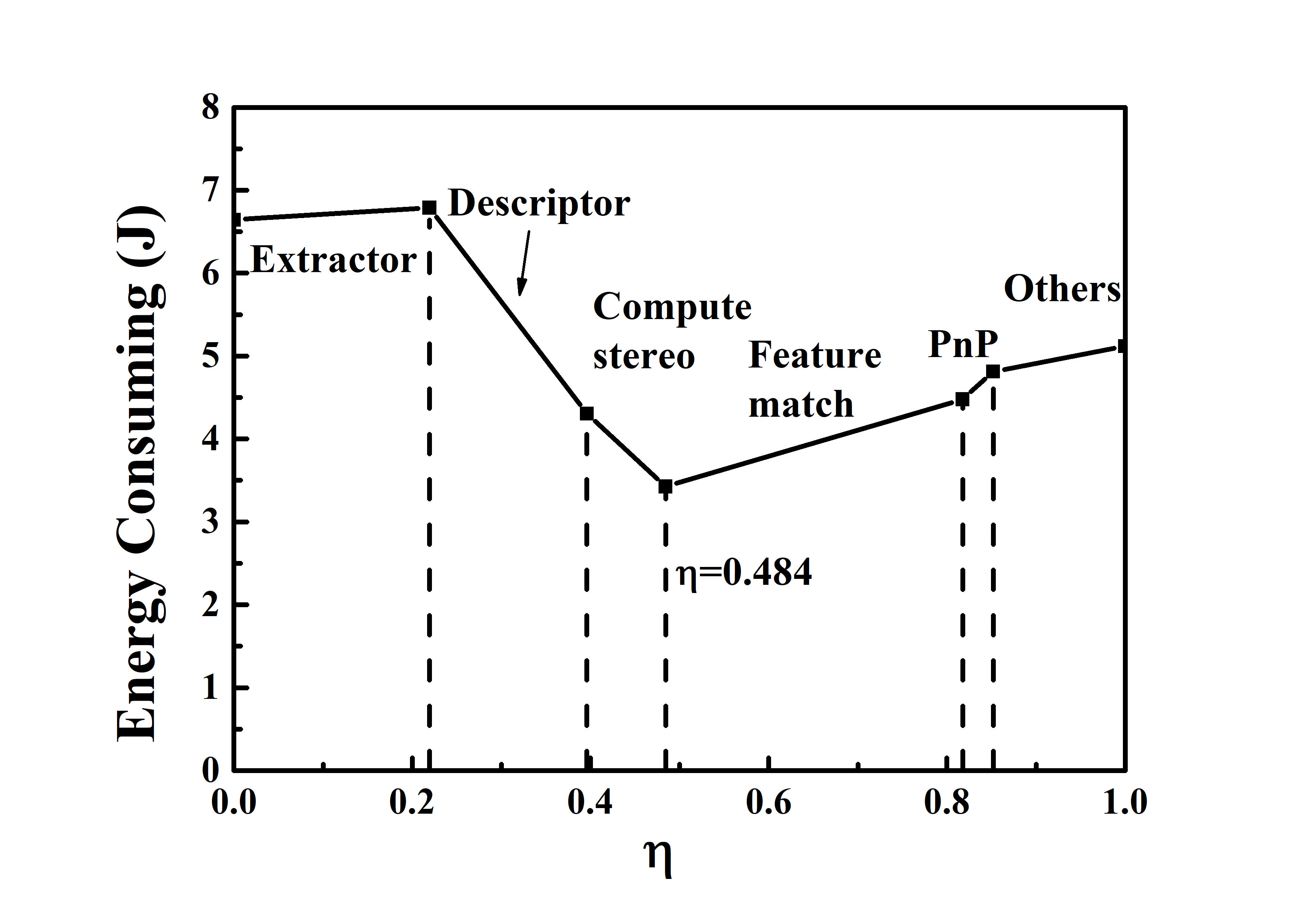}
  \caption{Relative energy consuming}
  \label{fig35}
\end{figure}

Based on (\ref{eq21}), the curve of the $E_{lk}$ can be seen in Fig. \ref{fig35}. From the curve we can clearly see, the best offloading point is $\eta=0.484$. That is to say, extractor, descriptor and compute stereo is calculated in the local device and other parts are offloaded to the cloud.

Now we have got the best value of $\eta$, then we design our experiment from following four aspects.

\subsubsection{Memory comparison}
\par
In this experiment, the key information extracted in each image, (which records the values of feature points, depths, descriptors, and time stamps), is stored in a YAML file. Comparing the original RGB image + Depth image with the YAML file including the feature point, the memory occupation can be seen in Fig.\ref{fig6a}.

As it is shown in Table \ref{tab3}, the frames catched by Kinect camera include RGB images and Depth images, which take up twice as much memory as YAML file.

\begin{table*}[tbp]
\centering
\caption{Memory size occupation of different files}
\label{table}

\begin{tabular}{|c|c|c|c|c|c|c|c|c|c|c|}
\hline
\qquad &1&2&3&4&5&6&7&8&9&10\\
\hline
YAML(kB)&
290.6&	290.6&	289.4&	290.6&	288.9&	291.2&	290.7&	290.1&	290.8&	290.5\\
\hline
RGB image(kB)&
506.5&	498&	498.8&	497&	501.3&	500.2&	507.7&	495.7&	484.7&	489.8\\
\hline
Depth image(kB)&
125.5&	124.9&	123.7&	123.5&	123.5&	122.3&	121.9&	121.2&	120.7&	120.9\\
\hline
RGB+Depth images(kB)&
632&	622.9&	622.5&	620.5&	624.8&	622.5&	629.6&	616.9&	605.4&	610.7\\
\hline
\end{tabular}
\label{tab3}
\end{table*}

\subsubsection{Transmission rate comparison}

\begin{figure}[htbp]
\centering
\subfloat[]{
\includegraphics[width=1.65in]{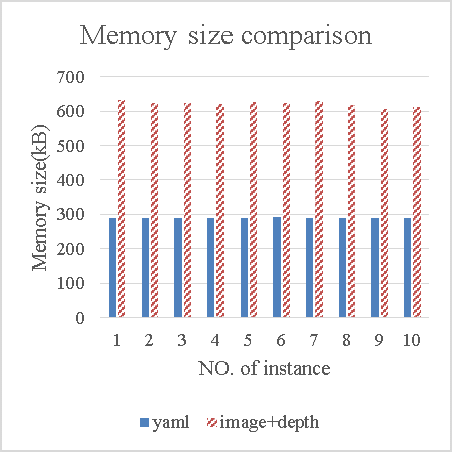}
\label{fig6a}}
\subfloat[]{
\includegraphics[width=1.65in]{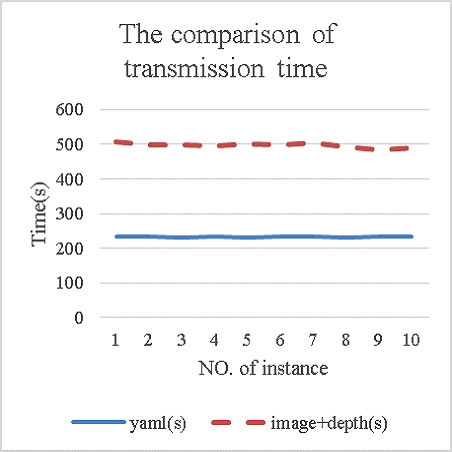}
\label{fig6b}}
\caption{(a) Memory size comparison; (b) Transmission time comparison}
\label{fig6}
\end{figure}

This paper assumed the bandwidth is 10Mbps. If there are 1000 frames needed to process, the time used by offloading all the images to the cloud (total offloading) and offloading the YAML file after extracting the feature point to the cloud (partial offloading) is compared in Fig.\ref{fig6b}. From the figure we can see that the transmission time is about 200s by using our partial offloading method, which is much less then total offloading.

\par

\subsubsection{Time of localization process comparison}
\par
\begin{figure}[htbp]
  \centering
  \includegraphics[width=3in]{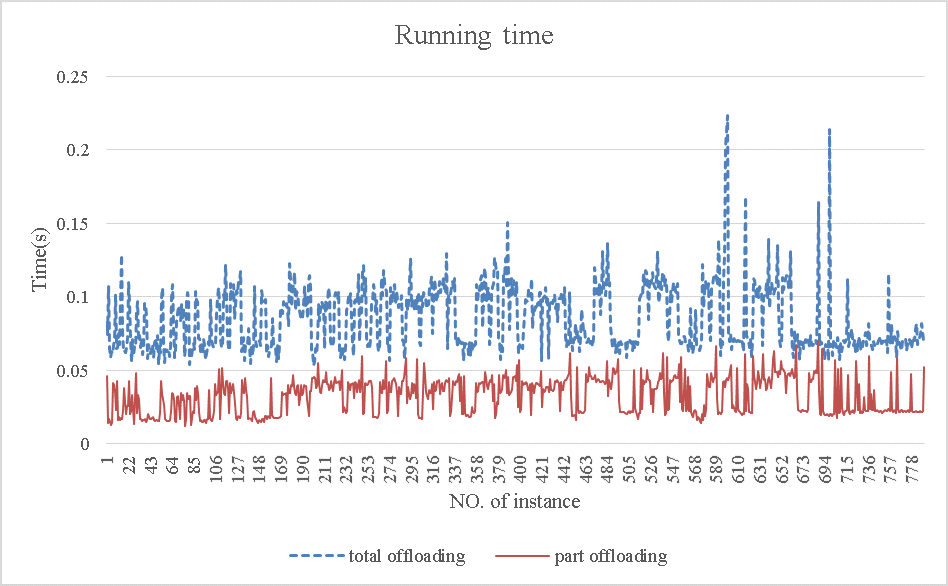}
  \caption{Running time comparison of the localization processing}
  \label{fig8}
\end{figure}

Since the tasks of feature point extraction are calculated by the embedded device, the speed of the localization processing is significantly improved, as shown in Fig.\ref{fig8}.

System average delay is shown in Table \ref{tab4} (for example there are 1000 frames). It can be seen that if offloaded all the tasks to the cloud have an average sum time of 578s. However, if we offloaded partial tasks to the cloud, the average total time cost is 327s. Actually, the average running time of the system is 282s, nearly one-half of the time of total offloading.

\begin{table}[htbp]
\centering
\caption{Running time comparison of the system }
\label{table}

\begin{tabular}{c|c|c|c|c|c}
\toprule[1pt]
\qquad &
robot&
trans&
cloud&
sum&
total\\

\hline
total offloading(s)&
0&	498&	80&	578&	578\\
\hline
partial offloading(s)&
50&	232&	45&	327&	282\\
\bottomrule[1pt]
\end{tabular}
\label{tab4}
\end{table}

\subsection{FS-HICP Algorithm}
\begin{figure}[htbp]
\centering
\subfloat[]{
\includegraphics[width=0.75in]{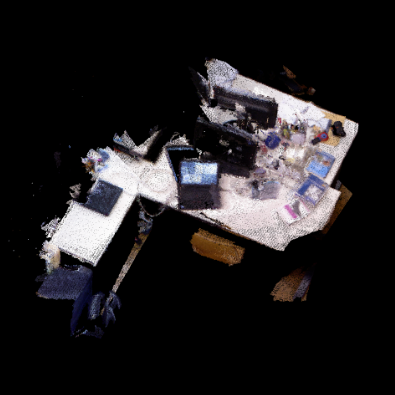}
\label{fig14a}}
\subfloat[]{
\includegraphics[width=0.75in]{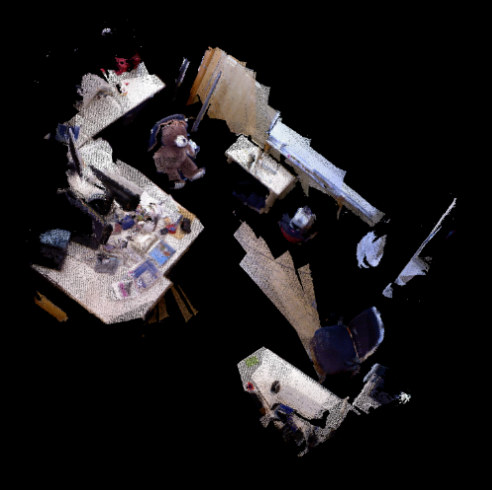}
\label{fig14b}}
\subfloat[]{
\includegraphics[width=0.75in]{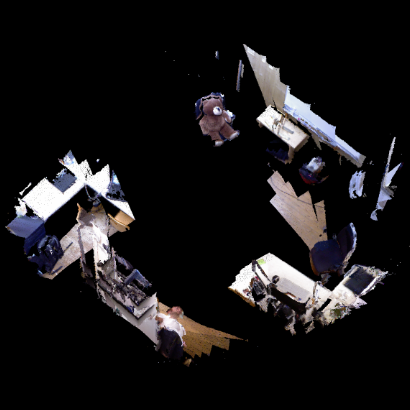}
\label{fig14c}}
\subfloat[]{
\includegraphics[width=0.75in]{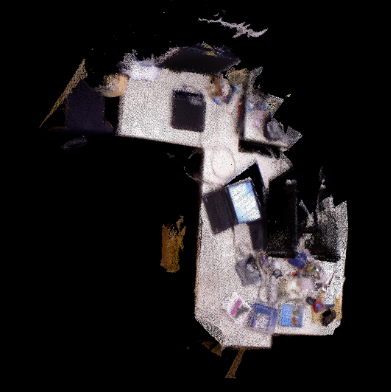}
\label{fig14d}}
\caption{Point cloud map obtained by different robots. (a) $Room_{1}$; (b) $Room_{2}$; (c) $Room_{3}$; (d) $Room_{4}$}
\label{fig14}
\end{figure}

In this experiment, there are four robots $R=\{R_{1},R_{2},R_{3},R_{4}\}$. And the local dense maps obtained by each robot can be expressed as $Room=\{Room_{1},Room_{2},Room_{3},Room_{4}\}$. The point cloud maps obtained by different robots are shown in Fig.\ref{fig14}.

\par

We design five experiments based on the $Room$ scene mentioned above to test the efficiency of the FS-HICP algorithm proposed in this paper.
\subsubsection{Point cloud voxel filtering}
\par
Since the partial room map obtained by each robot is very large, the first step of the algorithm is the point cloud map filtering. The number of the points in the map has dropped dramatically after filtering. The running time comparison of the ICP algorithm before and after filtering is shown in Table \ref{tab30}. It can be seen that after filtering, the running time of registration be greatly improved.

\begin{table*}[htbp]
\centering
\caption{Running time comparison of the mapping process }
\label{table}

\begin{tabular}{c|c|c|c|c|c|c|c|c|c|c}
\toprule[1pt]
\qquad &
1&2& 3 &4& 5& 6 &7 &8& 9& 10\\
\hline
time of classical ICP(ms)& 521269&	524367&	523680&	519856&	523154&	521236&	527189&	537834&	521238&	519807\\
\hline

time of FS-HICP(ms)&4540&	4569&	4493&	4526&	4498&	4446&	4487&	4533&	4496&	4510\\

\bottomrule[1pt]
\end{tabular}
\label{tab30}
\end{table*}

\subsubsection{FS-HICP algorithm time comparison}
\par
\begin{figure}[htbp]
\centering
\subfloat[]{
\includegraphics[width=1.65in]{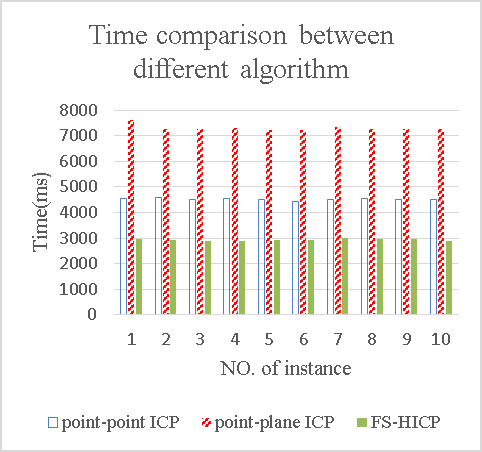}
\label{fig11a}}
\subfloat[]{
\includegraphics[width=1.67in]{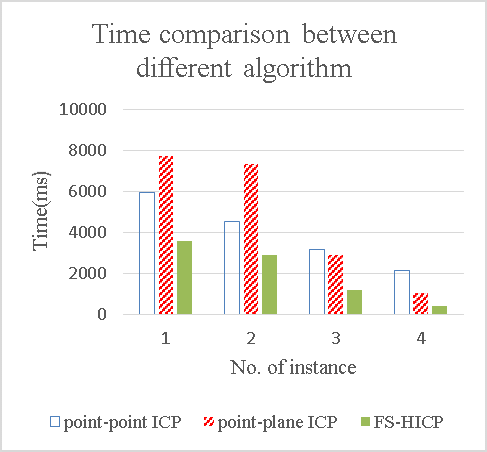}
\label{fig11b}}
\caption{Time comparison between different algorithm. (a) same data repeating multiple times; (b) four different data set repeat 10 times, in NO.1 and NO.2, the feature points are density, overlap area of No.2 is larger than NO.1, the data points in NO.3 and NO.4 are less than NO.1 and NO.2, feature points are relatively sparse.}
\label{fig11}
\end{figure}

In the indoor scenario discussed in this paper, two ICP algorithm strategies were tested: the basic \textit{point-point} ICP algorithm that comes from the point cloud library (PCL) and the \textit{point-plane} ICP algorithm used by Kinect fusion.  The time taken by the three algorithms is shown in Fig.\ref{fig11}. It was found that due to the large number of feature points in the indoor scene, point-plane method will generate a lot of errors. The convergence speed of FS-HICP method is much faster than the other two algorithms in four different situations.

\par
\subsubsection{Convergence speed comparison}
\par
\begin{figure}[htbp]
\centering
\subfloat[]{
\includegraphics[width=1.63in]{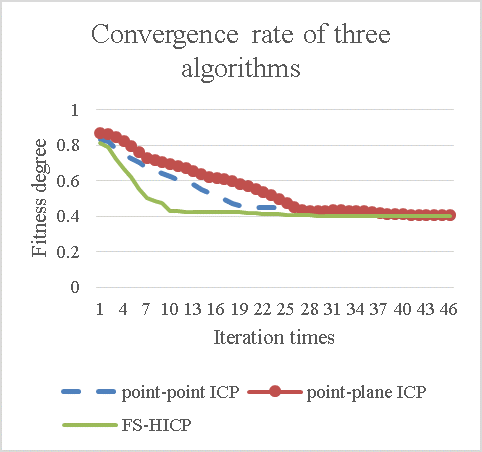}
\label{fig12a}}
\subfloat[]{
\includegraphics[width=1.67in]{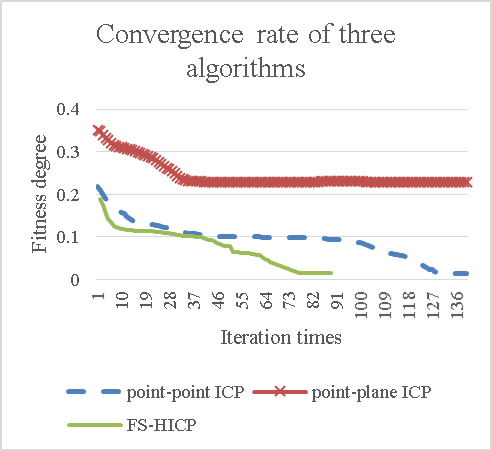}
\label{fig12b}}
\caption{Convergence rate comparison of three algorithms. (a) $Room_{1}$ and $Room_{4}$ registration; (b) $Room_{2}$ and $Room_{3}$ registration}
\label{fig12}
\end{figure}

This section analyzed two typical indoor scenes. In one case, $Room_{1}$ and $Room_{4}$ registration, there are fewer data points, but the overlapping parts are concentrated and the density of point clouds in the overlapping part is larger. In another case, $Room_{2}$ and $Room_{3}$ registration, there are more data points, including features of ground and walls, which have relatively fewer feature points including desktops, chairs and so on, in which the overlapping parts are scattered. Hence, it is difficult to achieve a right match in this case and the iteration time is larger than in first case. Fig.\ref{fig12a} shows that in relatively simple scene, FS-HICP algorithm can quickly reach convergence. In relatively complex scene, as shown in Fig.\ref{fig12b}, the point-plane algorithm cannot get a satisfactory result, and the convergence rate of point-point algorithm is slower than FS-HICP algorithm.

\par

\subsubsection{Error rate comparison}
This section experiments with the error rate of the algorithm as a function of iteration times. Firstly, the standard rotation and translation (transformation matrix) $T$ were given and then transformation matrix ${T}'$ was obtained after the iteration. The variance of ${T}'$ and $T$ was used to represent the error rate of the algorithm.

The standard transformation matrix $T$ is given in (\ref{eq4}).
\begin{equation} \label{eq4}
T=
\begin{bmatrix}
r_{00} &r_{01}  & r_{02} & t_{0}\\
r_{10} &r_{11}  &r_{12}  & t_{1}\\
r_{20} &r_{21}  &r_{22}  & t_{2}\\
 0& 0 & 0 & 1
\end{bmatrix}
\end{equation}

The transformation matrix ${T}'$ obtained after the iteration of the ICP algorithm is given in (\ref{eq5}).
\begin{equation} \label{eq5}
{T}'=
\begin{bmatrix}
{r_{00}}' &{r_{01}}'  & {r_{02}}' & {t_{0}}'\\
{r_{10}}' &{r_{11}}'  &{r_{12}}'  & {t_{1}}'\\
{r_{20}}' &{r_{21}}'  &{r_{22}}'  & {t_{2}}'\\
 0& 0 & 0 & 1
\end{bmatrix}
\end{equation}

The variance between the two matrixes is used as the criterion for error rate evaluation. It is expressed by $E_{rr}$, which can be seen in (\ref{eq6}).

\begin{equation} \label{eq6}
E_{rr}=\sum_{i=2}^{i=0}\sum_{j=2}^{j=0}(r_{ij}-{r}'_{ij})^{2}+\sum_{i=2}^{i=0}(t_{i}-{t}'_{i})^{2}-1
\end{equation}

\begin{figure*}[htbp]
\centering
\subfloat[]{
\includegraphics[width=2in]{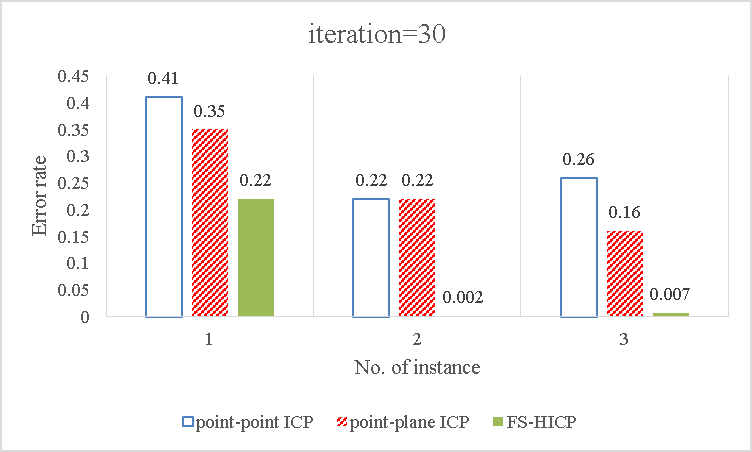}
\label{fig13a}}
\subfloat[]{
\includegraphics[width=2in]{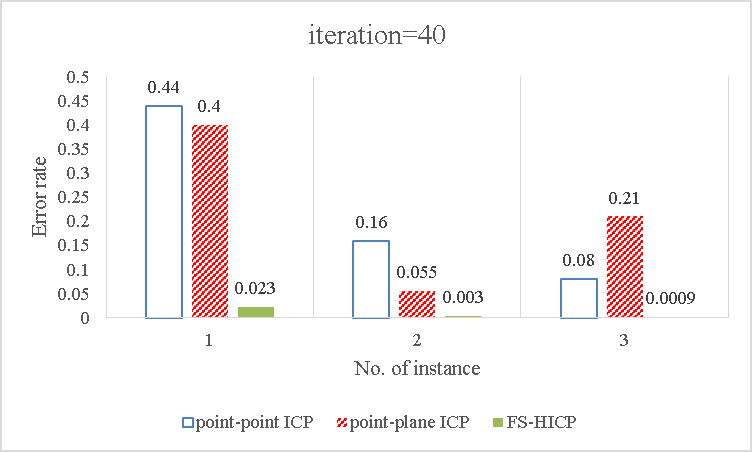}
\label{fig13b}}
\subfloat[]{
\includegraphics[width=2in]{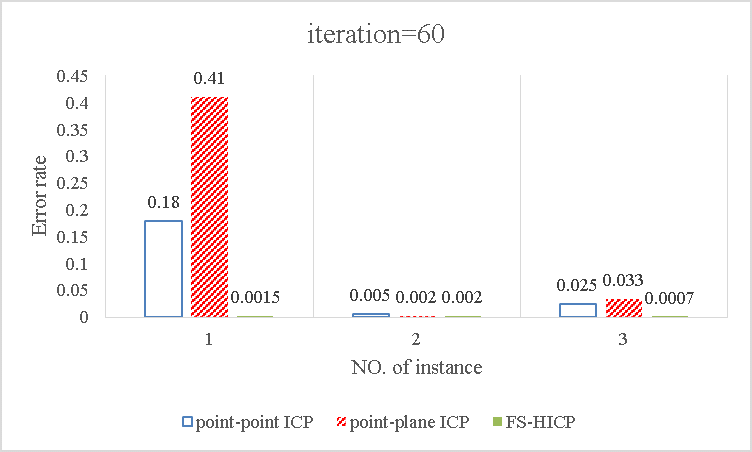}
\label{fig13c}}
\caption{Error rate of different algorithms. (a) iterating 30 times; (b) iterating 40 times; (c) iterating 60 times}
\label{fig13}
\end{figure*}

\begin{figure*}[htbp]
\centering
\subfloat[]{
\includegraphics[width=2.2in]{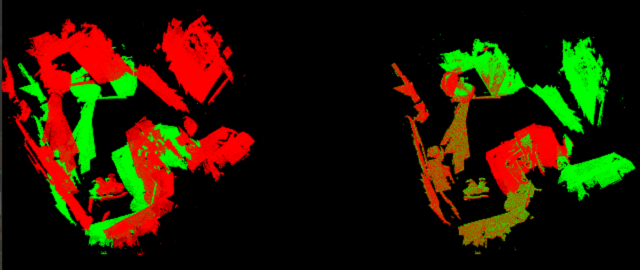}
\label{fig15a}}
\subfloat[]{
\includegraphics[width=2.2in]{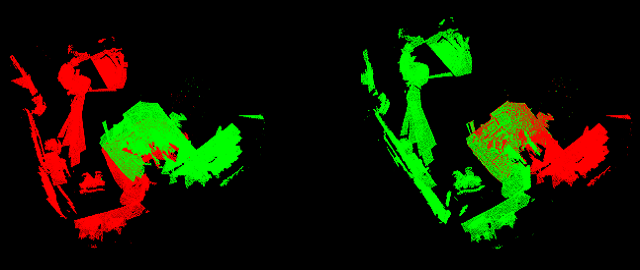}
\label{fig15b}}
\subfloat[]{
\includegraphics[width=2.2in]{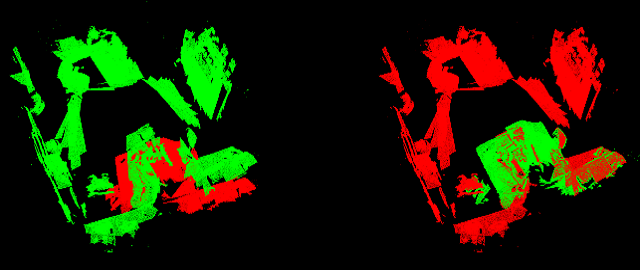}
\label{fig15c}}
\caption{Point cloud map registration. (a) $Room_{1}$ and $Room_{2}$ registration; (b) $Room_{2}$ and $Room_{3}$ registration; (c) $Room_{3}$ and $Room_{4}$ registration}
\label{fig15}
\end{figure*}

Fig.\ref{fig13} is the experiment result of error rate. We tested three different cases($Room_{2}$ and $Room_{3}$ registration; $Room_{1}$ and $Room_{2}$ registration; $Room_{3}$ and $Room_{4}$ registration). In each case, the three different algorithms iterated 30, 40, and 60 times separately. We can see, FS-HICP can get the same or even lower error rate than point-point algorithm. The registration processing of FP-HICP algorithm can be seen in Fig.\ref{fig15}.

\par

\subsubsection{Map building}
\begin{figure}[htbp]
\centering
\subfloat[]{
\includegraphics[width=1.3in]{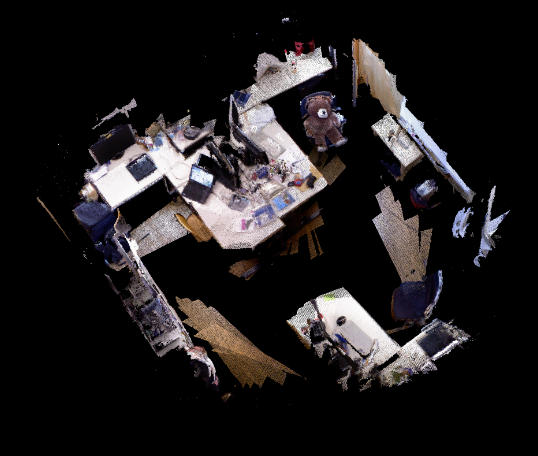}
\label{fig17a}}
\subfloat[]{
\includegraphics[width=1.3in]{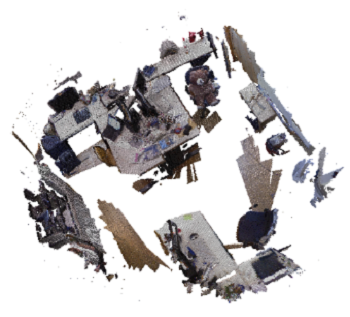}
\label{fig17b}}

\caption{Map building. (a) the global indoor dense point cloud map ; (b) the semi-dense octree map with resolution=16}
\label{fig17}
\end{figure}

The final global indoor point cloud map is shown in Fig.\ref{fig17a}. The map constructed by this method may be used for real-time navigation of robots. However, the point cloud maps occupy a lot of spaces and display too much unnecessary information. Additionally, it cannot reflect the transparency of space and unable to process moving objects and update maps in time. In this paper, the global point cloud maps were converted into octree maps for storage as shown in Fig.\ref{fig17b}. The memory space occupied by octree map is 4.9M while the memory space occupied by point cloud map is 11.4M. It turns out that octree map can save more space.
\vspace{-0.5cm}

\subsection{Practical Application}
In order to prove the validity of the whole system, in this section, we use two robotic cars equipped with Kinect cameras as experimental platforms and the two cars move independently in the scene.
\begin{figure*}[htbp]
\centering
\subfloat[]{
\includegraphics[width=2.25in]{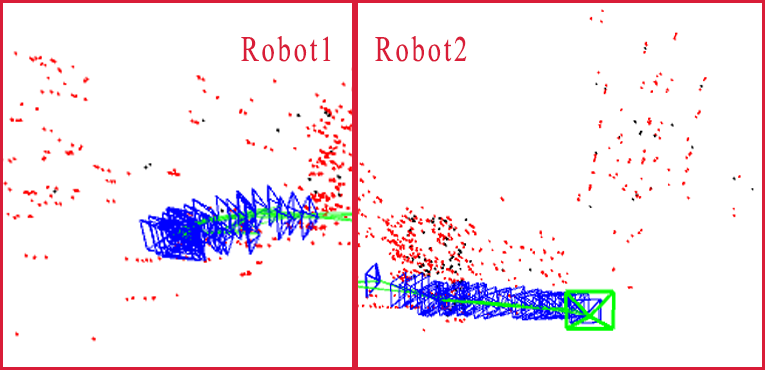}
\label{fig50a}}
\subfloat[]{
\includegraphics[width=2.6in]{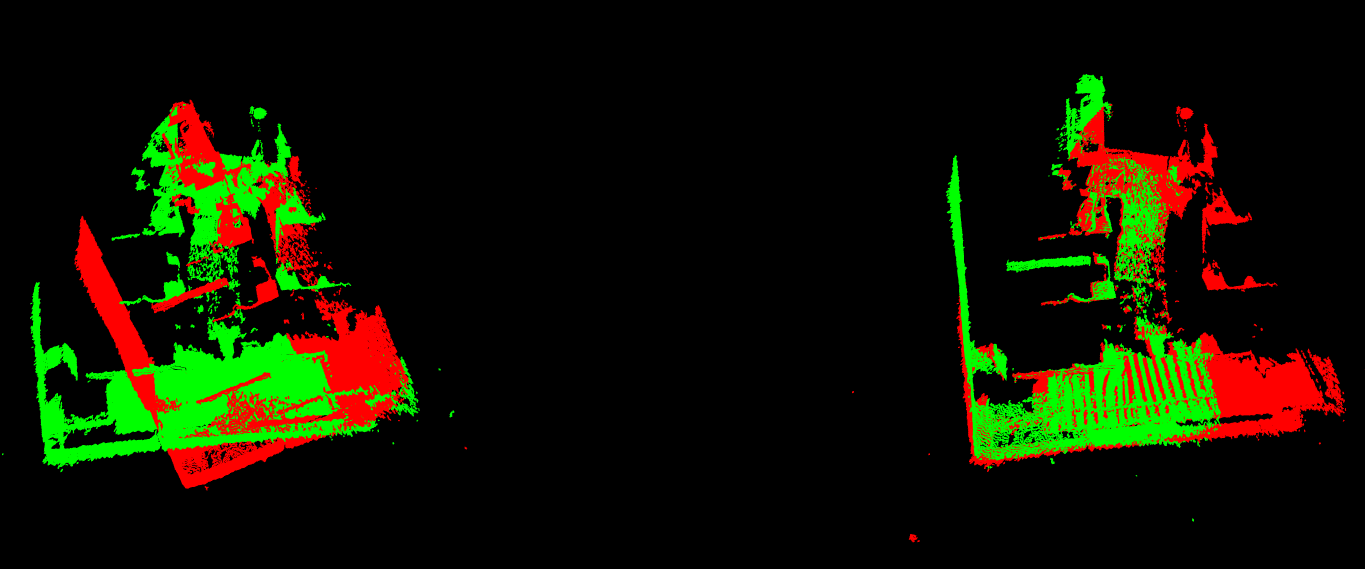}
\label{fig50b}}
\subfloat[]{
\includegraphics[width=1.3in]{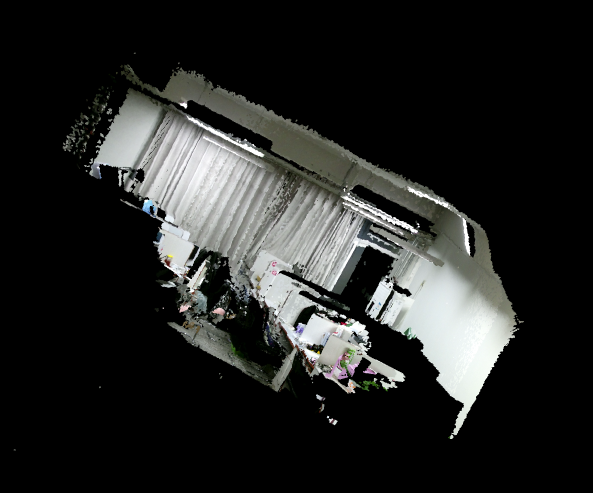}
\label{fig50c}}
\caption{Indoor scene application. (a) motion path of two robots; (b) $Room_{1}$ and $Room_{2}$ registration; (c) the global map of our lab}
\label{fig50}
\end{figure*}

\begin{figure*}[htbp]
\centering
\subfloat[]{
\includegraphics[width=2.3in]{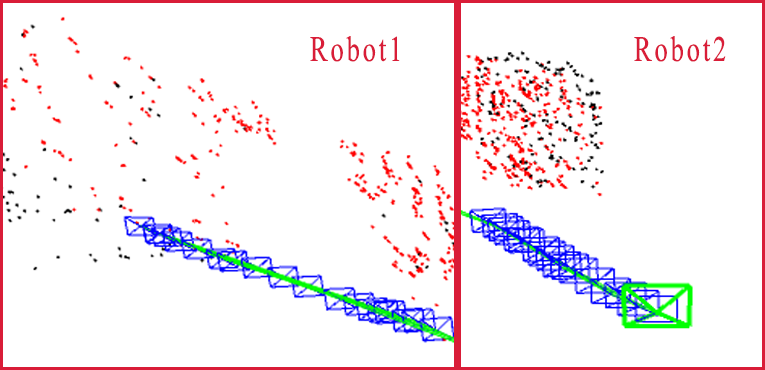}
\label{fig51a}}
\subfloat[]{
\includegraphics[width=2.7in]{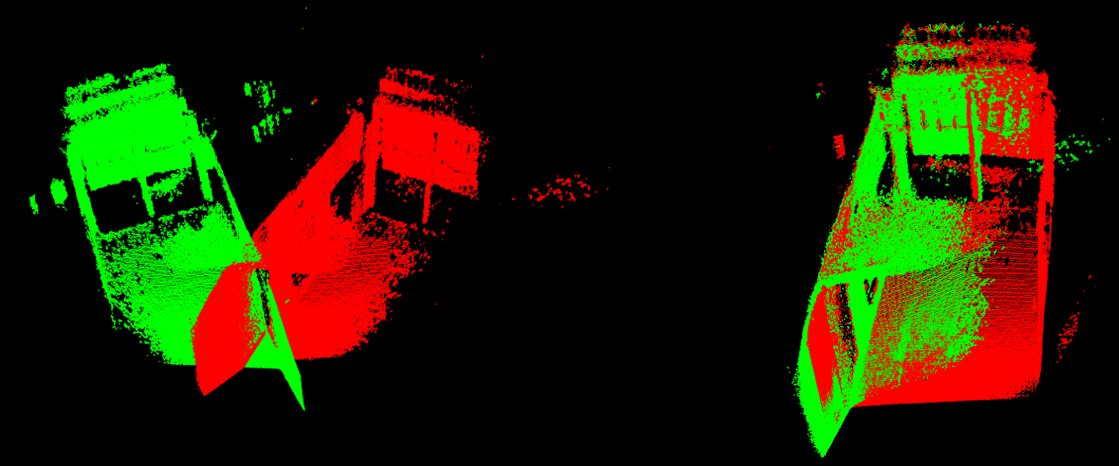}
\label{fig51b}}
\subfloat[]{
\includegraphics[width=1.1in]{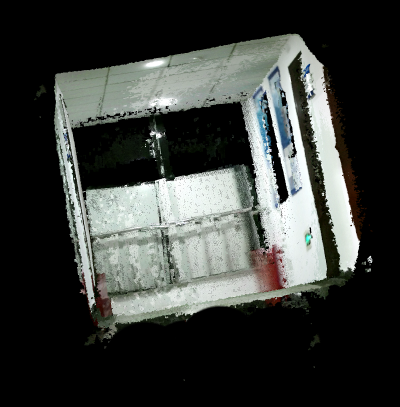}
\label{fig51c}}
\caption{Corridor scene application. (a) motion path of two robots; (b) $Corridor_{1}$ and $Corridor_{2}$ registration; (c) the global map of the corridor}
\label{fig51}
\end{figure*}

\begin{table*}[htbp]
\centering
\caption{Performance of the whole system }
\begin{tabular}{|c|c|c|c|c|c|c|c|}

\toprule[1pt]
                                &                                  & \multicolumn{2}{c|}{$E_{lk}$} & $E_{localization}$        & $E_{mapping}$              & $E_{system}$                 & $T_{total}(s)$             \\ \hline
\multirow{4}{*}{Indoor scene}   & \multirow{2}{*}{our system}      & $E_{l1}$        & 221.2        & \multirow{2}{*}{537.2} & \multirow{2}{*}{3.15}  & \multirow{2}{*}{540.35} & \multirow{2}{*}{291.45}  \\ \cline{3-4}
                                &                                  & $E_{l2}$        & 316          &                        &                        &                         &                        \\ \cline{2-8}
                                & \multirow{2}{*}{original system} & $E_{l1}$        & 1019.2       & \multirow{2}{*}{2475.2} & \multirow{2}{*}{6.93}  & \multirow{2}{*}{2482.13} & \multirow{2}{*}{1176.8} \\ \cline{3-4}
                                &                                  & $E_{l2}$       & 1456          &                        &                        &                         &                        \\ \bottomrule[1pt] \toprule[1pt]
\multirow{4}{*}{Corridor scene} & \multirow{2}{*}{our system}      & $E_{l1}$        & 95.3        & \multirow{2}{*}{249.3}  & \multirow{2}{*}{4.35}  & \multirow{2}{*}{253.65}  & \multirow{2}{*}{154.1}  \\ \cline{3-4}
                                &                                  & $E_{l2}$        & 154        &                        &                        &                         &                        \\ \cline{2-8}
                                & \multirow{2}{*}{original system} & $E_{l1}$        & 436.8        & \multirow{2}{*}{1164.8}  & \multirow{2}{*}{10.95} & \multirow{2}{*}{1175.75}  & \multirow{2}{*}{610.8} \\ \cline{3-4}
                                &                                  & $E_{l2}$        & 728        &                        &                        &                         &                        \\ \bottomrule[1pt]
\end{tabular}
\label{tab55}
\end{table*}
\subsubsection{Indoor scene}
Our lab is used as an indoor scene. In this scenario, the number of feature points is large and the environment is relatively complicated. The motion path of the two robots can be seen in Fig.\ref{fig50a}. The blue boxes in the figure represent the keyframes and the green lines are the motion path of the robots. Global keypoints are the red points in the figure while local keypoints are represented by black points. It is assumed that the local map got by $Robot_{1}$ is $Room_{1}$ and the local map obtained by $Robot_{2}$ is $Room_{2}$. The registration of $Room_{1}$ and $Room_{2}$ can be seen in Fig.\ref{fig50b} and the point cloud map after splicing is shown in Fig.\ref{fig50c}. Since the frames got by Kinect cameras have distortion error, which are not as accurate as the frames in TUM data set, we can see there is a little dislocation in Fig.\ref{fig50c}, but it does not affect the overall map.

\subsubsection{Corridor scene}
We used corridor out of our lab as another scene where the number of feature points is small and the environment is relatively simple. The motion path of two robots can be seen in Fig.\ref{fig51a}. It is assumed the local map obtained by $Robot_{1}$ is $Corridor_{1}$ and the local map obtained by $Robot_{2}$ is $Corridor_{2}$. The registration of $Corridor_{1}$ and $Corridor_{2}$ can be seen in Fig.\ref{fig51b}. Furthermore the point cloud map after splicing is shown in Fig.\ref{fig51c}.

\subsubsection{Performance of the whole system}
We tested the whole system performance based on the experiments mentioned in the two sections above. The energy consumed by robot $R_{1}$ is $E_{l1}$ while $E_{l2}$ represents the energy consumed by robot$R_{2}$. $E_{localization}$ can be calculated by equation (\ref{eq52}). Since the pixel of image obtained by our Kinect is $1280\times 960$, the memory occupied by images are much larger than YAML file obtained by our partial offloading method. As for the original system, most of the energy is consumed on the network transmission. From the Table \ref{tab55} we can see that the energy consumption and time consumed by our algorithm are nearly one quarter of the original algorithm. No matter in indoor scene or in corridor scene, our method can greatly reduce the system energy consumption and time cost.

\section{Conclusion and future work}
Firstly, this paper proposes a multi-robot visual SLAM partial computing offloading strategy where the best offloading point is given to reduce the energy consumption and time cost of the whole visual SLAM system so that tasks that could not be achieved on the robot embedded device become possible. Secondly, an improved point cloud map registration algorithm (FS-HICP algorithm) suitable for indoor scene is proposed, which improves the convergence speed dramatically. Finally, we build the global dense point cloud map and semi-dense octree map. In some special environments, such as fire, underwater, underground situations etc., the method proposed in this paper can have a wider application.

In the future, we will continue our research of multi-robot visual SLAM, aiming at solving the problem of communications between multiple robots, realizing multi-robot path planning, using the method of deep learning to achieve semantic SLAM and so on.

\appendices
\section{}
\subsection{The concept used in FS-HICP algorithm}
\textit{Fitness score}: Fitness score is the sum of the squared distances from source to target. It is supposed that the point $p_{i}=(x_{i},y_{i},z_{i})$ is in the source point cloud and point ${p_{i}}'=({x_{i}}',{y_{i}}',{z_{i}}')$ is in the target point cloud. Their Euclidean distance is described in (\ref{eq7}) and the fitness score is described in (\ref{eq8}).

\begin{equation} \label{eq7}
\begin{split}
d(p_{i},{p_{i}}')=\left \| p_{i}-{p_{i}}' \right \|=\\
\sqrt{(x_{i}-{x_{i}}')^{2}+(y_{i}-{y_{i}}')^{2}+(z_{i}-{z_{i}}')^{2}}
\end{split}
\end{equation}

\begin{equation} \label{eq8}
fitnessscore=\frac{\sum_{i=1}^{n}d(p_{i},{p_{i}}')}{n}
\end{equation}

\textit{Maximum Response Distance}: It reflects the maximum distance threshold between two corresponding points in the source and target. If the distance between two points is greater than this threshold, these points will be ignored during the alignment process.

\textit{Simulated Annealing algorithm}: Simulated annealing is a randomized algorithm used to find the extremum value of a function. In this paper, the simulated annealing algorithm is used to prevent the algorithm from being trapped into local optimality.

\textit{Layered Filtering}: The idea of layered filtering is used to accelerate the iteration. In this paper, it was found that the result of two layers is the best. The layered filtering first uses the voxel filter of the larger grid to complete the coarse matching. After the matching accuracy reaches the given constraint, the voxel filter with the smaller grid is used to achieve further precision matching.

\subsection{YAML file structure}
The YAML file structure used in partial offloading model is shown in Table \ref{tab1}.
\begin{table}[htbp]
\caption{The structure of YAML file}
\label{table}

\begin{tabular}{|p{220pt}|}
\hline
\%YAML:1.0 \par ---

KeyPoint:
\par
- [ 265... -1 ]
\par
...
\par
- [ 3.4756860351562500e+02... -1 ]
\par
Depth: [ 1.61939991e+00, ..., 1.04139996e+00 ]
\par
Descriptor: !!opencv-matrix
\par
rows: 1008
\par
cols: 32
\par
dt: u
\par
data: [ 38,8,..., 218 ]
\par
Time: 1.3050311021753039e+09\\

\hline
\end{tabular}
\label{tab1}
\end{table}

\bibliographystyle{IEEEtran}
\nocite{*}
\bibliography{reference}





\end{document}